%% file: cut20.tex
\documentclass[journal]{IEEEtran}

\usepackage{cite}

\ifCLASSINFOpdf
\usepackage[pdftex]{graphicx}
\else
\fi
\usepackage{amsmath,amssymb,amsfonts}
\usepackage[caption=false]{subfig}
\usepackage{threeparttable}
\usepackage{multirow}
\usepackage{cite}
\usepackage{microtype}
\usepackage{dblfloatfix}
\usepackage{tikz}

\usepackage{pifont}
\let\oldding\ding
\renewcommand{\ding}[2][1]{\scalebox{#1}{\oldding{#2}}}
\newcommand{\ineq}[1]{\footnotesize$#1$\normalsize}{}

\newcommand{\prior}{\text{{SpiNeMap}}}{}

\begin{document}
\bstctlcite{IEEEexample:BSTcontrol}
\title{Reliability-Performance Trade-offs \\in Neuromorphic Computing
}

\author{\IEEEauthorblockN{Twisha Titirsha and Anup Das}\\
\IEEEauthorblockA{{Electrical and Computer Engineering}
{Drexel University}, Philadelphia, PA, USA\\
Email: \{tt624,anup.das\}@drexel.edu}
}

\maketitle

\begin{abstract}
\input{sections/abstract}
\end{abstract}


\section{Introduction}\label{sec:introduction}
\input{sections/introduction}


\section{Background}\label{sec:background}
\input{sections/background}

\section{Endurance-Performance Trade-offs in Neuromorphic Hardware}\label{sec:problem_formulation}
\input{sections/problem}

\section{Mapping Explorations}\label{sec:prior}
\input{sections/prior}

\section{Evaluation}\label{sec:evaluation}
\input{sections/evaluation}

\section{Conclusions}\label{sec:conclusions}
\input{sections/conclusion}

\section*{Acknowledgment}
This work is supported by the National Science Foundation Award CCF-1937419 (RTML: Small: Design of System Software to Facilitate Real-Time Neuromorphic Computing).





\IEEEtriggeratref{24}
\bibliographystyle{IEEEtran}
\bibliography{Commands,Disco,External}

\end{document}

%% file: sections/abstract.tex
Neuromorphic architectures built with Non-Volatile Memory (NVM) can significantly improve the energy efficiency of machine learning tasks designed with Spiking Neural Networks (SNNs).
A major source of voltage drop in a crossbar of these architectures are the parasitic components on the crossbar's bitlines and wordlines, which are deliberately made longer to achieve lower cost-per-bit.
We observe that the parasitic voltage drops create a significant asymmetry in programming speed and reliability of NVM cells in a crossbar.
Specifically, NVM cells that are on shorter current paths are faster to program but have lower endurance than those on longer current paths, and vice versa. This asymmetry in neuromorphic architectures create reliability-performance trade-offs, which can be exploited efficiently using SNN mapping techniques.
In this work, we demonstrate such trade-offs using a previously-proposed SNN mapping technique with 10 workloads from contemporary machine learning tasks for a state-of-the art neuromoorphic hardware.

%% file: sections/introduction.tex
Spiking Neural Networks (SNNs) are emerging machine learning models with spike-based computation and bio-inspired learning algorithms. Event-driven neuromorphic hardware such as TrueNorth~\cite{truenorth}, Loihi~\cite{loihi}, and DYNAP-SE~\cite{dynapse} implements biological neurons and synapses to execute SNN-based machine learning tasks in an energy-efficient manner. This makes neuromorphic hardware suitable for energy-constrained platforms such as the embedded systems~\cite{lee2016introduction} and edge devices of the Internet-of-Things (IoTs)~\cite{shi2016edge}.

A neuromorphic hardware is implemented as a tile-based architecture, the tiles are interconnected using a shared interconnect such as the Network-on-Chip (NoC)~\cite{benini2002networks} and Segmented Bus~\cite{sbGLSVLSI}. Each tile consists of a crossbar, which can implement a fixed number of neurons and synapses. A crossbar in a neuromorphic hardware is an \ineq{n\times n} organization, with \ineq{n} bitlines (columns) and \ineq{n} worklines (rows). A silicon neuron is mapped along each wordline of a crossbar, while a synaptic cell is placed at the cross-section of each bitline and wordline using an access device such as a transistor or a diode~\cite{catthoor2018very}.

Recently, Non-Volatile Memory (NVM) such as Phase-Change Memory (PCM), Oxide-based Resistive RAM (OxRRAM), and and Spin-Transfer Torque Magnetic or Spin-Orbit-Torque RAM (STT- and SoT-MRAM) are used as synaptic cells to increase integration density and reduce energy consumption of crossbars in neuromorphic hardware~\cite{Burr2017,Mallik2017,wijesinghe2018all}.

A major source of voltage drops in a crossbar are the parasitic resistance and capacitance on its bitlines and wordlines, which are deliberately made longer to achieve lower cost-per-bit. In fact, for a PCM-based crossbar, each neuron is approximately 18x the size of a PCM cell~\cite{indiveri2003low}. To amortize this large size, systems designers implement larger crossbars, e.g., \ineq{128\times 128} for DYNAP-SE and \ineq{256\times 256} for TrueNorth. For such large crossbar sizes, the current on the longest path in a crossbar becomes significantly lower than the current on its shortest path for the same spike voltage generated from a neuron and the same conductance programmed on the enabled synaptic cell in these paths.\footnote{The length of a current path in a crossbar is measured in terms of the number of parasitic components that are encountered on the path.}

Current asymmetry leads to a difference in performance and reliability of NVM cells. Higher current through an NVM cell can lead to faster programming of the cell. This means that NVM cells on shorter current paths are faster to access and program. 
However, NVMs also have limited endurance, ranging from
\ineq{10^5} (for Flash) to \ineq{10^{10}} (for OxRRAM), with PCM somewhere in between
(\ineq{\approx 10^{7}}). An NVM cell's endurance is strongly dependent on the programming current. We build the case for PCM, where the conductance change is induced by Joule heating of the chalcogenide material in the cell. The endurance of the material depends on the self-heating temperature, which is dependent on the programming current. Therefore, the NVM cells on shorter current paths have higher self-heating temperature, and therefore lower endurances.

In recent years, many approaches are proposed to map SNNs to neuromorphic hardware. This includes the performance-oriented SNN mapping technique of~\cite{dfsynthesizer,balaji2020ESL}, the dataflow-based mapping technique of~\cite{das2018dataflow,balaji2019frameworkISVLSI}, the energy-aware mapping technique of~\cite{pycarl,spinemap,das2018mapping}, the circuit aging-aware mapping technique of~\cite{frameworkCAL,reneu,NeuromorphicLR,lcpc}, and the run-time SNN mapping technique of~\cite{balaji2020run}.
Unfortunately, none of these approaches exploit the reliability and performance trade-offs of NVM cells in neuromorphic computing. 
In this paper, we take one such mapping approach -- \prior{}, and show the significant variations in endurance and speed during its mapping explorations.

The remainder of this paper is organized as follows. We provide a background of PCM and neuromorphic architectures in Section~\ref{sec:background}. Next, we formulate the endurance-access speed trade-offs for a single PCM cell and integrate such trade-offs at the crossbar-level in Section~\ref{sec:problem_formulation}. Next, we discuss the mapping exploration of \prior{} in Section~\ref{sec:prior}. We present our evaluation in Section~\ref{sec:evaluation} and conclusion in Section~\ref{sec:conclusions}.

%% file: sections/background.tex
In this section, we discuss the background on Phase-Change Memory (PCM) to aid the understanding of the trade-offs in neuromorphic computing. We also provide a discussion on machine learning approaches using SNNs, and how such approaches can be mapped to hardware consisting of PCM cells organized into crossbars.

A PCM cell is built with chalcogenide alloy, e.g., Ge${}_2$Sb${}_2$Te${}_5$ (GST)~\cite{ovshinsky1968reversible}, and is connected to a bitline and a wordline using an access device. 
The GST alloy can either be in an amorphous (high resistance) state, or in one of the partially crystallized (low resistance) states.
PCM is recently explored as scalable DRAM alternative for conventional computing~\cite{LeeISCA2009,QureshiISCA09,palp,mneme,datacon,hebe}. This work explores PCM for neuromorphic computing.
For such computing architectures, the weight of a synaptic connection is programmed as conductance of a PCM cell by driving current and inducing Joule heating in the cell.

In many machine learning approaches such as online learning using Spike-Timing Dependent Plasticity (STDP)~\cite{kheradpisheh2018stdp}, one-shot learning~\cite{fei2006one}, life-long learning~\cite{allred2019stimulating}, and reinforcement learning~\cite{shim2017biologically}, it is necessary to update synaptic weights based on the input excitation. To facilitate such synaptic updates, a PCM cell's state must be switched by driving current through it using the spikes generated from neurons. However, frequent switching of a PCM cell's state may lead to endurance issues, where the cell fails to be programmed correctly, leading to a degradation of machine learning performance.
Furthermore, a key requirement in such online learning use-cases is their real-time performance, i.e., the weight updates must be completed within a small time interval.

\begin{figure}[h!]
	\begin{center}
		\includegraphics[width=0.99\columnwidth]{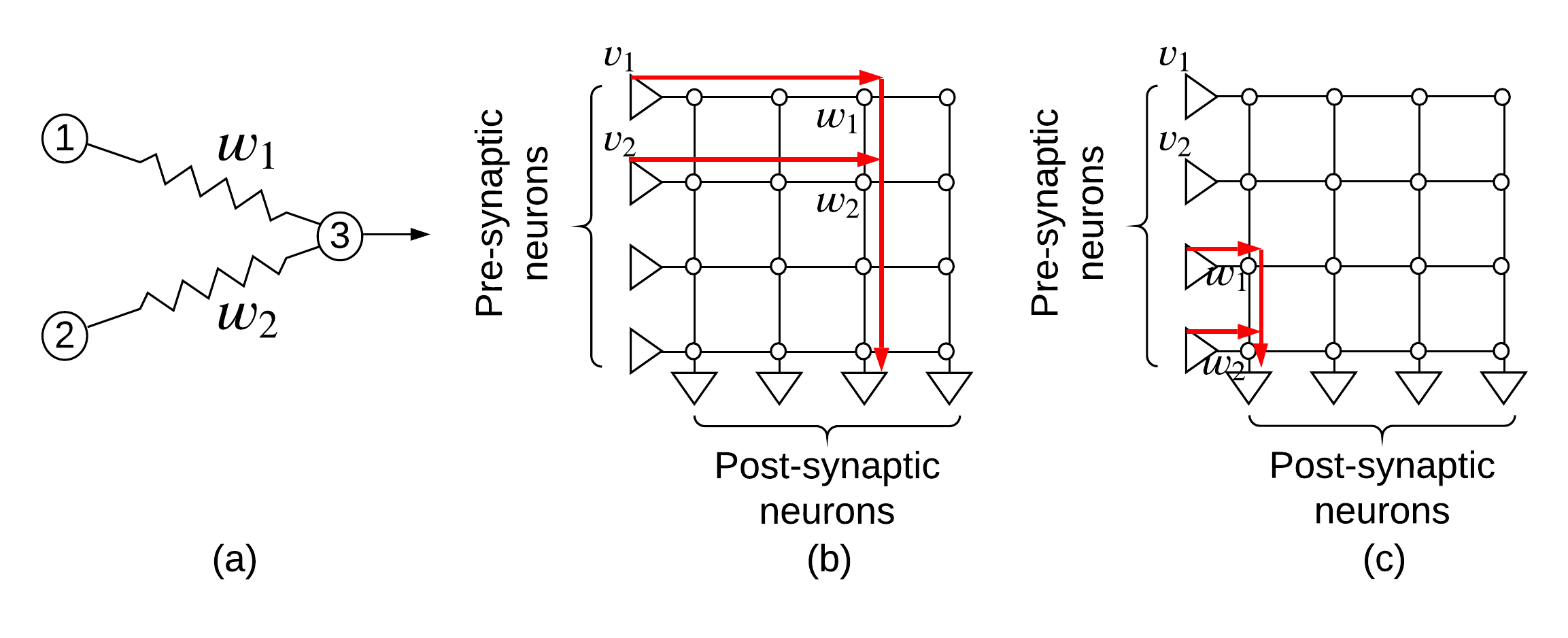}
		\vspace{-10pt}
		\caption{(a) A simple spiking neural network, (b) Mapping of the network to a crossbar, (c) A different mapping of the network to the hardware.}
		\label{fig:mapping}
	\end{center}
\end{figure}

To understand the workload-dependent performance and endurance trade-offs associated with PCM cells in a neuromorphic architecture, Figure~\ref{fig:mapping}a shows a simple SNN with two input and one output neurons. Figure~\ref{fig:mapping}b illustrates the mapping of this SNN to a crossbar. As seen from this figure, the synaptic weights \ineq{w_1} and \ineq{w_2} are programmed as conductances. The spike voltages are multiplied with the conductances to generate current, which gets integrated along the columns. The current strength guides the update of conductance of the PCM cell enabled along the current path. Clearly, the weight update frequency depends on the spikes generated from the hardware neurons mapped along the rows of a crossbar. The latter depends on how neurons and synapses of a machine learning model, e.g., Figure~\ref{fig:mapping}a are mapped to the corresponding resources on a crossbar. To elaborate on this, Figure~\ref{fig:mapping}c illustrates the utilization of a different set of PCM cells to realize the SNN of Figure~\ref{fig:mapping}a. If the PCM cells in a crossbar have different performance and endurance characteristics (which we demonstrate in this work), then the mapping of neurons and synapses of a machine learning model plays a critical role in system-level performance and reliability. 

%% file: sections/problem.tex
We formulate the endurance-performance trade-offs for a single PCM cell. To establish the relationship, 
we consider the GST material of a PCM cell to be in a crystalline state. The amorphization process, i.e., the crystalline-to-amorphous state transition involves driving a very high current through the cell for a short duration.
This high current raises the temperature of the GST material through Joule heating in the heater attached to the GST, which transitions the material to its amorphous state. The crystalline fraction (\ineq{V_c}) is computed using the Johnson-Mehl-Avrami (JMA) equation~\cite{Avrami1941} as
\begin{equation}
    \label{eq:vc}
    \footnotesize V_c=\text{exp}\left[-\alpha \times\frac{(T_{SH}-T_{amb})}{T_m}\times t\right],
\end{equation}
where \ineq{t} is the time, \ineq{T_m} is the melting temperature of the GST material, and \ineq{\alpha} is a fitting constant.
The exponential decay of \ineq{V_c} in Equation~\ref{eq:vc} implies that, higher the self-heating temperature (\ineq{T_{SH}}), faster is the reduction of the crystalline volume, i.e., faster is the amorphization process.

The self-heating temperature is related to the square of programming current (\ineq{I_\text{prog}}) as
\begin{equation}
    \label{eq:tsh}
    \footnotesize T_{SH}=k\cdot I_\text{prog}^2
\end{equation}
where \ineq{k} is a constant.

From Equations~\ref{eq:vc} and \ref{eq:tsh} we can conclude that higher the programming current, higher is the self-heating temperature, and hence, faster is the programming of the cell.

However, with increase in self-heating temperature, the endurance of a PCM cell reduces. Using the phenomenological endurance model~\cite{Strukov2016}, endurance of a PCM cell can be expressed as
\begin{equation}
    \label{eq:endurance}
    \footnotesize \text{Endurance}\approx \text{exp}\left(\frac{\gamma}{T_{SH}}\right),
\end{equation}
where \ineq{\gamma} is  a fitting parameter.

From Equations~\ref{eq:tsh} and \ref{eq:endurance}, we conclude that higher the programming current, higher is the self-heating temperature and therefore, lower is the endurance.

\begin{figure}[h!]
	\begin{center}
		\includegraphics[width=0.89\columnwidth]{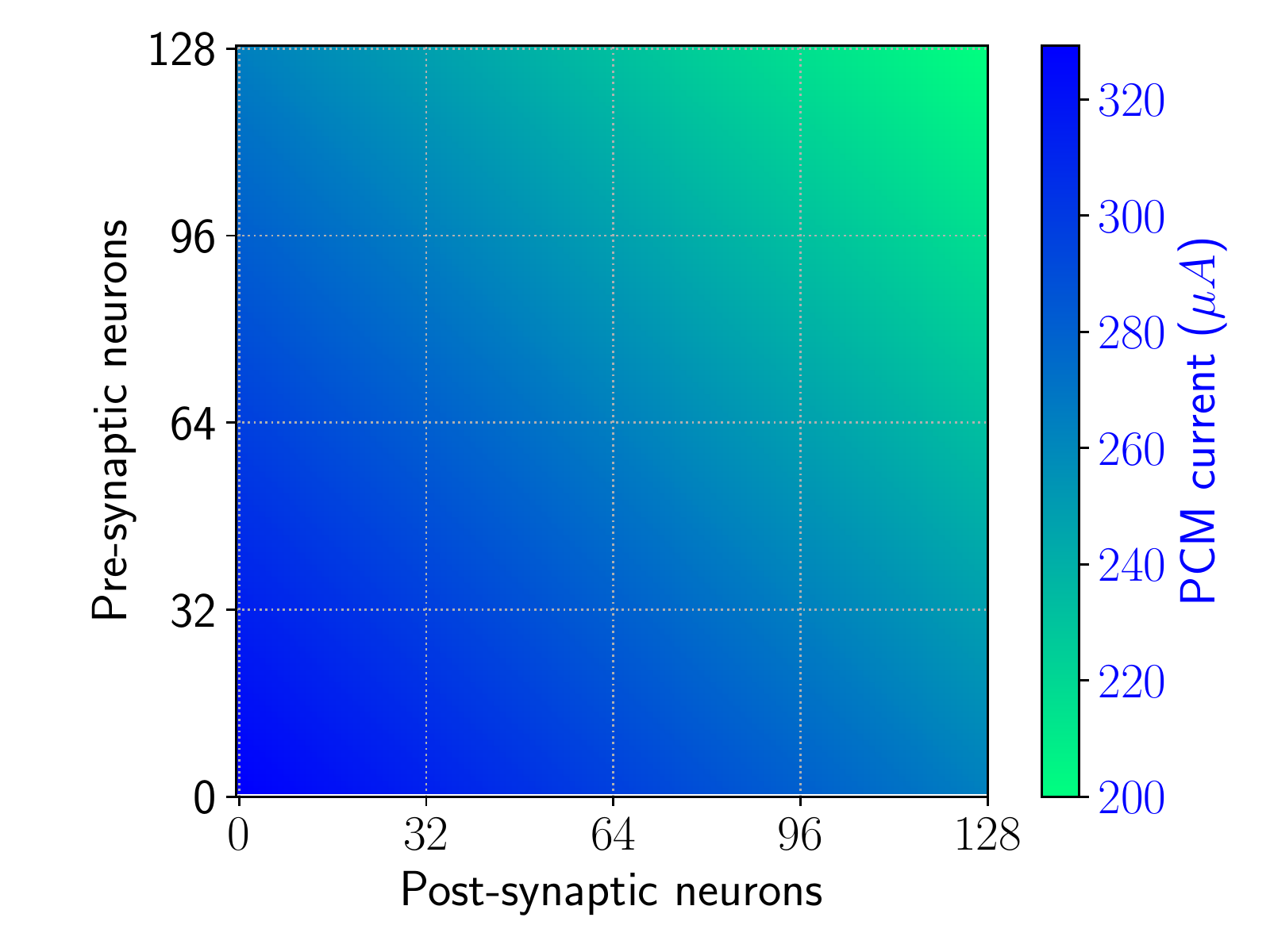}
		\vspace{-10pt}
		\caption{Current map in a 128x128 crossbar.}
		\label{fig:current_map}
	\end{center}
\end{figure}

Figure~\ref{fig:current_map} shows the current through the PCM cells in a 128x128 PCM crossbar. This current variation is due to the difference in the length of current paths from pre-synaptic neurons to post-synaptic neurons in the crossbar, where the length of a current path is measured in terms of the number of parasitic elements on the path. These current values are obtained for a 65nm technology node and at 300K temperature corner. As can be clearly seen from the figure, current through PCM cells on the top-right corner of the crossbar is lower than through PCM cells located at the bottom-left corner. Therefore, cells at the top-right corner are slower to program and have higher endurance, while those at the bottom-left corner are faster to program and have lower endurance. Table~\ref{tab:tradeoff} summarizes these findings.

\begin{table}[h!]
	\renewcommand{\arraystretch}{1.4}
	\setlength{\tabcolsep}{1.4pt}
	\caption{Summary of performance-endurance trade-offs.}
	\label{tab:tradeoff}
	\centering
	{\fontsize{6}{9}\selectfont
		\begin{tabular}{|c|c|c|}
			\hline
			\textbf{Location} & \textbf{Performance} & \textbf{Endurance}\\
			\hline
			Top-right corner & \textcolor{red}{Low} & \textcolor{blue}{High}\\
			Bottom-left corner & \textcolor{blue}{High} & \textcolor{red}{Low}\\
            \hline
	    \end{tabular}
	 }
\end{table}

%% file: sections/prior.tex
In this section, we present the mapping exploration of \prior{}~\cite{spinemap} and show the performance-endurance trade-offs that are obtained during its design-space exploration.

\prior{} uses an instance of the Particle Swarm Optimization (PSO)~\cite{kennedy2010particle}, a meta-heuristic approach to map neurons and synapses to the hardware. To this end, \prior{} first partitions a spiking neural network based application into clusters, where each cluster can fit onto the resources of a crossbar. The clusters are then mapped to the crossbars using the PSO. In general, PSO finds the optimum solution to a fitness function $F$. Each solution is represented as a particle in the swarm. Each particle has a velocity with which it moves in the search space to find the optimum solution. During the movement, a particle updates its position and velocity according to its own experience (closeness to the optimum) and also experience of its neighbors. We introduce the following notations for PSO.

\begin{footnotesize}
 	\begin{align}
 	\label{eq:pso_defn}
 	D &= \text{dimensions of the search space}\\
 	n_p &= \text{number of particles in the swarm}\nonumber \\
 	\mathbf{\Theta} = \{\mathbf{\theta}_l\in\mathbb{R}^{D}\}_{l=0}^{n_p-1} &= \text{positions of particles in the swarm}\nonumber \\
 	\mathbf{V} = \{\mathbf{v}_l\in\mathbb{R}^{D}\}_{l=0}^{n_p-1} &= \text{velocity of particles in the swarm}\nonumber 
 	\end{align}
 \end{footnotesize}

Position and velocity updates are performed according to the following equation.

\begin{figure}[h!]
	\centering
	\centerline{\includegraphics[width=0.99\columnwidth]{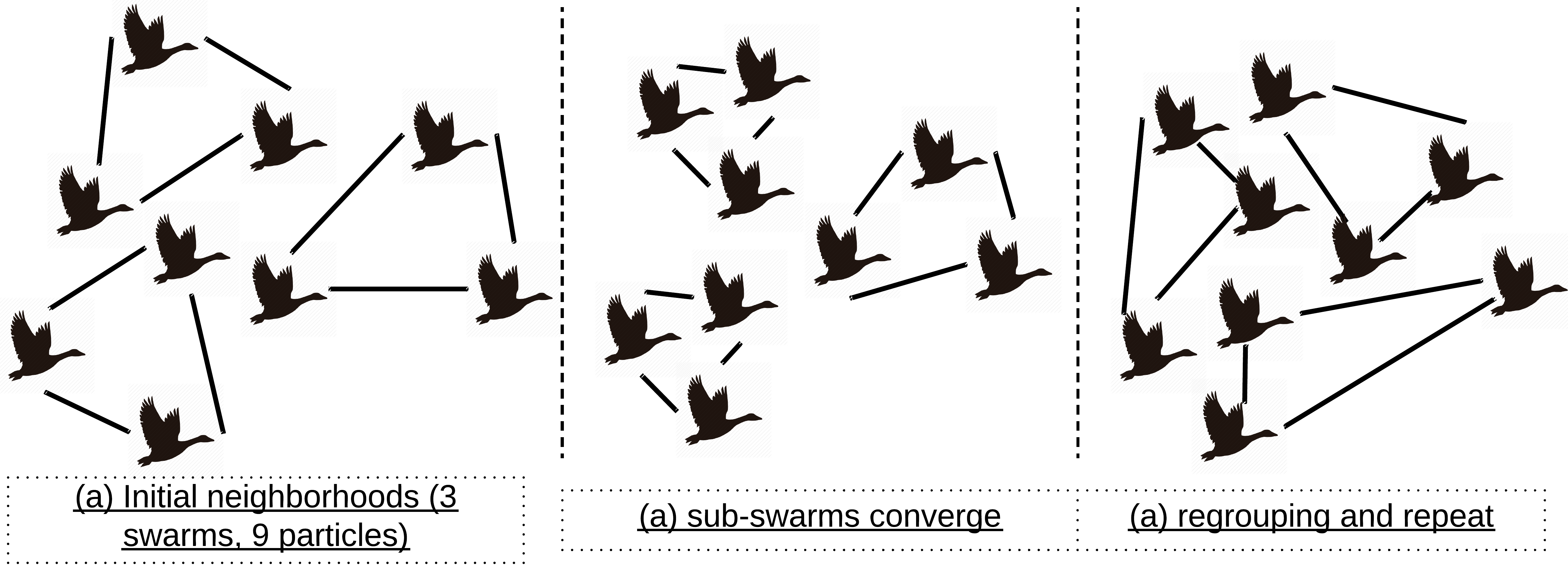}}
	\caption{Illustrating the iterative steps of PSO.}
	\label{fig:pso_steps}
\end{figure}

\begin{footnotesize}
	\begin{align}
	\label{eq:pos_vel_update}
	\mathbf{\Theta}(t+1) &= \mathbf{\Theta}(t) + \mathbf{V}(t+1)\\
	\mathbf{V}(t+1) &= \mathbf{V}(t) + \varphi_1\cdot\Big(P_{\text{best}}-\mathbf{\Theta}(t)\Big) + \varphi_2\cdot\Big(G_{\text{best}}-\mathbf{\Theta}(t)\Big)\nonumber
	\end{align}
\end{footnotesize}
\normalsize where $t$ is the iteration number, $\varphi_1,\varphi_2$ are constants and $P_{\text{best}}$ (and $G_{\text{best}}$) is the particles own (and neighbors) experience. In Figure \ref{fig:pso_steps}, we illustrate the iterative approach to find an optimal solution using PSO. The PSO algorithm starts with an initial neighborhood of swarms. In this example, we illustrate 3 swarms, each with 3 particles (see Figure \ref{fig:pso_steps}a). Each particle jumps to a new location with a velocity determined as a function of local best (within swarms), and global best. This continues until the sub-swarms converge (see Figure \ref{fig:pso_steps}b). In the third step, the swarm regroups and the position and velocity update steps are repeated (see Figure \ref{fig:pso_steps}c). We continue these iterations until a predefined convergence criteria is reached.

\prior{} uses PSO to minimize the number of spikes communicated on the global interconnect, which leads to a reduction in the energy consumption.

\begin{figure}[h!]
	\centering
	\centerline{\includegraphics[width=0.99\columnwidth]{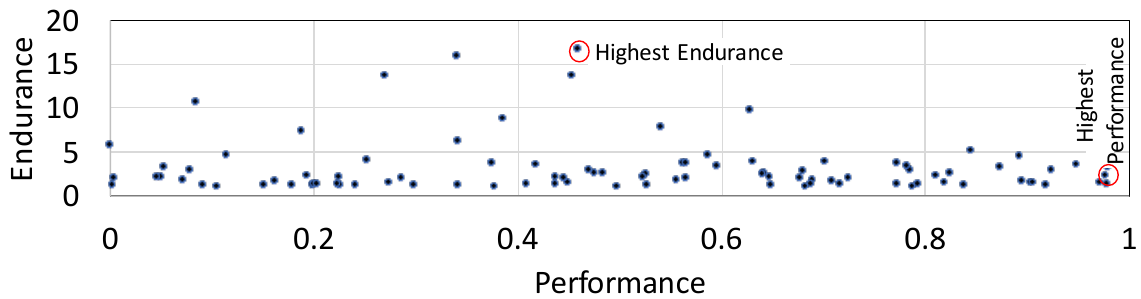}}
	\caption{Performance-endurance trade-offs using \prior{}.}
	\label{fig:tradeoff}
\end{figure}

Figure~\ref{fig:tradeoff} illustrates the performance-endurance trade-offs obtained during the mapping exploration of \prior{}. The figure plots the performance and endurance obtained for different design solutions generated during the design-space exploration using the PSO. The figure also shows two solutions -- one with highest endurance and one with the highest performance. We note that the highest performance mapping is generated by DFSynthesizer~\cite{dfsynthesizer}, which maps neurons and synapses to hadware, minimizing the execution time of applications.

%% file: sections/evaluation.tex
\subsection{Evaluation Framework}
We evaluated 10 machine learning applications that are representative of three most commonly used neural network classes --- convolutional neural network (CNN), multi-layer perceptron (MLP), and recurrent neural network (RNN).
These applications are 
1) LeNet~\cite{lenet} based handwritten digit recognition with \ineq{28 \times 28} images of handwritten digits from the MNIST dataset \cite{deng2012mnist};
2) AlexNet~\cite{alexnet} for Imagenet classification~\cite{deng2009imagenet};
3) VGG16~\cite{vgg16}, also for Imagenet classification~\cite{deng2009imagenet};
4) ECG-based heart-beat classification (HeartClass)~\cite{das2018heartbeat,HeartClassJolpe} using electrocardiogram (ECG) data from the Physionet database~\cite{moody2001physionet};
5) {multi-layer perceptron (MLP)-based handwritten digit recognition} (MLP-MNIST)~\cite{Diehl2015} using the MNIST database;
6) {edge detection} (EdgeDet)~\cite{carlsim} on $64 \times 64$ images using difference-of-Gaussian; 
7) {image smoothing} (ImgSmooth)~\cite{carlsim} on $64 \times 64$ images; 
8) {heart-rate estimation} (HeartEstm)~\cite{HeartEstmNN} using ECG data;
9) RNN-based predictive visual pursuit (VisualPursuit)~\cite{Kashyap2018}; and
10) recurrent digit recognition (R-DigitRecog)~\cite{Diehl2015}.
Table~\ref{tab:apps} summarizes the topology, the number of neurons and synapses of these applications, and their baseline accuracy.
To demonstrate the trade-offs, we enable STDP-based weight updates~\cite{kheradpisheh2018stdp} in each of these applications.\footnote{Spike-Timing Dependent Plasticity (STDP)~\cite{dan2004spike} is a learning mechanism in SNNs, where the synaptic weight between a pre- and a post-synaptic neuron is updated based on the timing of pre-synaptic inputs relative to the post-synaptic spike.} But our approach is not limited to STDP. 

\begin{table}[h!]
	\renewcommand{\arraystretch}{0.8}
	\setlength{\tabcolsep}{2pt}
	\caption{Evaluated applications.}
	\label{tab:apps}
	\centering
	\begin{threeparttable}
	{\fontsize{6}{10}\selectfont
		\begin{tabular}{cc|ccl|c}
			\hline
			\textbf{Class} & \textbf{Applications} & \textbf{Synapses} & \textbf{Neurons} & \textbf{Topology} & \textbf{Accuracy}\\
			\hline
			\multirow{4}{*}{CNN} & LeNet~\cite{lenet} & 282,936 & 20,602 & CNN & 85.1\%\\
			& AlexNet~\cite{alexnet} & 38,730,222 & 230,443 & CNN & 90.7\%\\
			& VGG16~\cite{vgg16} & 99,080,704 & 554,059 & CNN & 69.8 \%\\
			& HeartClass~\cite{HeartClassJolpe} & 1,049,249 & 153,730 & CNN & 63.7\%\\
			\hline
			\multirow{3}{*}{MLP} & DigitRecogMLP & 79,400 & 884 & FeedForward (784, 100, 10) & 91.6\%\\
			& EdgeDet \cite{carlsim} & 114,057 &  6,120 & FeedForward (4096, 1024, 1024, 1024) & 100\%\\
			& ImgSmooth \cite{carlsim} & 9,025 & 4,096 & FeedForward (4096, 1024) & 100\%\\
			\hline
 			\multirow{3}{*}{RNN} & HeartEstm \cite{HeartEstmNN} & 66,406 & 166 & Recurrent Reservoir & 100\%\\
 			& VisualPursuit \cite{Kashyap2018} & 163,880 & 205 & Recurrent Reservoir & 47.3\%\\
 			& R-DigitRecog \cite{Diehl2015} & 11,442 & 567 & Recurrent Reservoir & 83.6\%\\
			\hline
	\end{tabular}}
	\end{threeparttable}
\end{table}

We model the DYNAP-SE neuromorphic hardware~\cite{dynapse} with the following configurations.

\begin{itemize}
    \item A tiled array of 4 tiles, each with a 128x128 crossbar. There are 65,536 memristors per crossbar.
    \item Spikes are digitized and communicated between cores through a mesh routing network using the Address Event Representation (AER) protocol.
    \item Each synaptic element is a PCM-based memristor. 
\end{itemize}

Table \ref{tab:hw_parameters} reports the hardware parameters of DYNAP-SE.

\begin{table}[h!]
    \caption{Major simulation parameters extracted from \cite{dynapse}.}
	\label{tab:hw_parameters}
	\centering
	{\fontsize{6}{10}\selectfont
		\begin{tabular}{lp{5cm}}
			\hline
			Neuron technology & 65nm CMOS\\
			\hline
			Synapse technology & PCM\\
			\hline
			Supply voltage & 1.0V\\
			\hline
			Energy per spike & 50pJ at 30Hz spike frequency\\
			\hline
			Energy per routing & 147pJ\\
			\hline
			Switch bandwidth & 1.8G. Events/s\\
			\hline
	\end{tabular}}
\end{table}

We evaluate the following metrics.
\begin{itemize}
    \item \textbf{Performance:} This is the time it takes to execute an application on the hardware model.
    \item \textbf{Effective lifetime:} This is the minimum effective lifetime of all PCM cells in the hardware. The \emph{effective lifetime} (\ineq{\mathcal{L}_{i,j}}), defined for the PCM cell connecting the \ineq{i^\text{th}} pre-synaptic neuron with \ineq{j^\text{th}} post-synaptic neuron in a memristive crossbar as
\begin{equation}
    \label{eq:normalized_lifetime}
    \footnotesize \mathcal{L}_{i,j} =  \mathcal{E}_{i,j}/a_{i,j}, 
\end{equation}
where \ineq{a_{i,j}} is the number of spikes propagating through the PCM cell in a given SNN workload and \ineq{\mathcal{E}_{i,j}} is its endurance.
\end{itemize}

\subsection{Performance}
\begin{figure}[h!]
	\centering
	\centerline{\includegraphics[width=0.99\columnwidth]{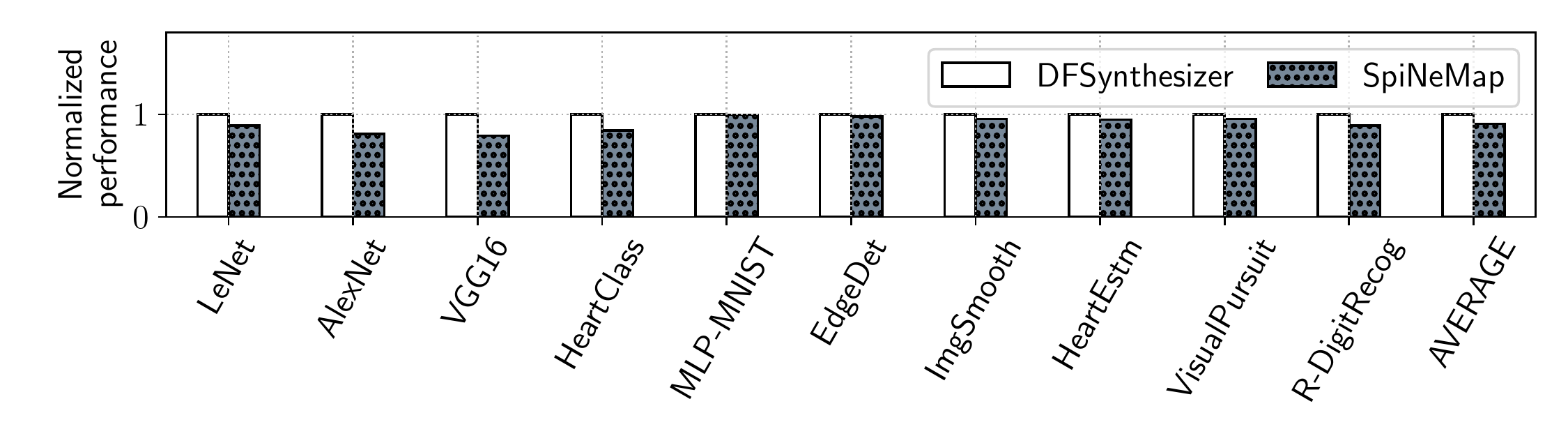}}
	\caption{Performance normalized to DFSynthesizer.}
	\label{fig:performance}
\end{figure}

Figure~\ref{fig:performance} compares the performance of DFSynthesizer, a performance-oriented technique to map SNNs to neuromorphic hardware and \prior{}, which minimizes the number of spikes on the shared interconnect. We observe that compared to DFSynthesizer, the performance using \prior{} is an average 10\% lower for these applications.

\subsection{Effective Lifetime}
\begin{figure}[h!]
	\centering
	\centerline{\includegraphics[width=0.99\columnwidth]{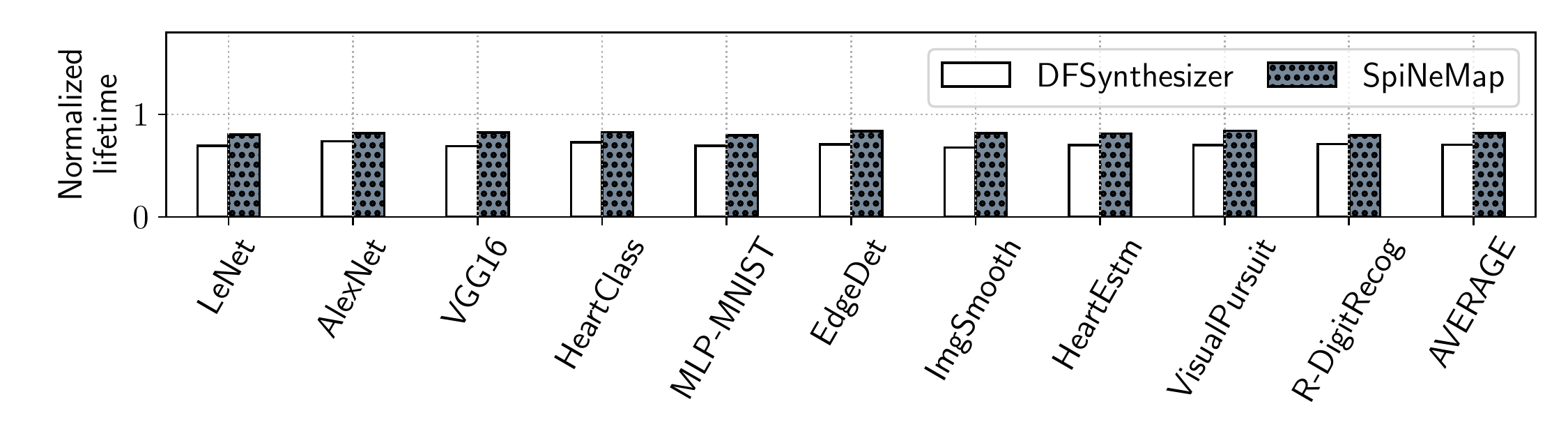}}
	\caption{Lifetime normalized to mapping with highest lifetime.}
	\label{fig:lifetime}
\end{figure}

Figure~\ref{fig:lifetime} plots the normalized lifetime of DFSynthesizer and \prior{} for the evaluated applications. Lifetime results are normalized to the lifetime obtained using the mapping that generates the highest effective lifetime (see Figure~\ref{fig:tradeoff}). We observe that lifetime using the mapping of DFSynthesizer is on average 30\% lower, while that using \prior{} is 19\% lower than the highest lifetime.

%% file: sections/conclusion.tex
In this work, we show the trade-offs between performance and lifetime of neuromorphic hardware with PCM-based crossbars. Specifically, we show that in a PCM-based crossbar, the PCM cells that are located on the bottom-left corner are faster to access but have lower lifetime than PCM cells on the top-right corner, which are slower but have higher lifetime. Existing SNN-mapping techniques do not explore this trade-offs in mapping neurons and synapses to hardware. The design space exploration of these mapping techniques often select mapping that generate high performance or optimize for energy consumption. Therefore, the lifetime obtained using these techniques is significantly lower than the highest lifetime. A possible future direction is therefore, to explore the trade-offs during the design-space exploration. This will enable generating SNN mapping that are balanced in terms of lifetime, performance, and energy consumption.

%% file: cut20.bbl
\begin{thebibliography}{10}
\providecommand{\url}[1]{#1}
\csname url@samestyle\endcsname
\providecommand{\newblock}{\relax}
\providecommand{\bibinfo}[2]{#2}
\providecommand{\BIBentrySTDinterwordspacing}{\spaceskip=0pt\relax}
\providecommand{\BIBentryALTinterwordstretchfactor}{4}
\providecommand{\BIBentryALTinterwordspacing}{\spaceskip=\fontdimen2\font plus
\BIBentryALTinterwordstretchfactor\fontdimen3\font minus
  \fontdimen4\font\relax}
\providecommand{\BIBforeignlanguage}[2]{{%
\expandafter\ifx\csname l@#1\endcsname\relax
\typeout{** WARNING: IEEEtran.bst: No hyphenation pattern has been}%
\typeout{** loaded for the language `#1'. Using the pattern for}%
\typeout{** the default language instead.}%
\else
\language=\csname l@#1\endcsname
\fi
#2}}
\providecommand{\BIBdecl}{\relax}
\BIBdecl

\bibitem{truenorth}
M.~V. Debole \emph{et~al.}, ``{TrueNorth: Accelerating from zero to 64 million
  neurons in 10 years},'' \emph{Computer}, 2019.

\bibitem{loihi}
M.~Davies \emph{et~al.}, ``{Loihi: A neuromorphic manycore processor with
  on-chip learning},'' \emph{IEEE Micro}, 2018.

\bibitem{dynapse}
S.~Moradi \emph{et~al.}, ``{A scalable multicore architecture with
  heterogeneous memory structures for dynamic neuromorphic asynchronous
  processors (DYNAPs)},'' \emph{TBCAS}, 2017.

\bibitem{lee2016introduction}
E.~A. Lee \emph{et~al.}, \emph{Introduction to embedded systems: A
  cyber-physical systems approach}.\hskip 1em plus 0.5em minus 0.4em\relax Mit
  Press, 2016.

\bibitem{shi2016edge}
W.~Shi \emph{et~al.}, ``Edge computing: Vision and challenges,'' \emph{IOTJ},
  2016.

\bibitem{benini2002networks}
L.~Benini \emph{et~al.}, ``Networks on chip: A new paradigm for systems on chip
  design,'' in \emph{DATE}, 2002.

\bibitem{sbGLSVLSI}
A.~Balaji \emph{et~al.}, ``Exploration of segmented bus as scalable global
  interconnect for neuromorphic computing,'' in \emph{GLSVLSI}, 2019.

\bibitem{catthoor2018very}
F.~Catthoor \emph{et~al.}, ``Very large-scale neuromorphic systems for
  biological signal processing,'' in \emph{CMOS Circuits for Biological Sensing
  and Processing}, 2018.

\bibitem{Burr2017}
G.~W. Burr \emph{et~al.}, ``{Neuromorphic computing using non-volatile
  memory},'' \emph{Advances in Physics: X}, 2017.

\bibitem{Mallik2017}
A.~Mallik \emph{et~al.}, ``{Design-technology co-optimization for OxRRAM-based
  synaptic processing unit},'' in \emph{VLSIT}, 2017.

\bibitem{wijesinghe2018all}
P.~Wijesinghe \emph{et~al.}, ``An all-memristor deep spiking neural computing
  system: A step toward realizing the low-power stochastic brain,''
  \emph{TETCI}, 2018.

\bibitem{indiveri2003low}
G.~Indiveri, ``{A low-power adaptive integrate-and-fire neuron circuit},'' in
  \emph{ISCAS}, 2003.

\bibitem{dfsynthesizer}
S.~Song \emph{et~al.}, ``{Compiling spiking neural networks to neuromorphic
  hardware},'' in \emph{LCTES}, 2020.

\bibitem{balaji2020ESL}
A.~Balaji \emph{et~al.}, ``Enabling resource-aware mapping of spiking neural
  networks via spatial decomposition,'' \emph{Embedded Systems Letters}, 2020.

\bibitem{das2018dataflow}
A.~Das \emph{et~al.}, ``Dataflow-based mapping of spiking neural networks on
  neuromorphic hardware,'' in \emph{GLSVLSI}, 2018.

\bibitem{balaji2019frameworkISVLSI}
A.~Balaji \emph{et~al.}, ``A framework for the analysis of
  throughput-constraints of snns on neuromorphic hardware,'' in \emph{ISVLSI},
  2019.

\bibitem{pycarl}
A.~Balaji \emph{et~al.}, ``{PyCARL: A PyNN interface for hardware-software
  co-simulation of spiking neural network},'' in \emph{IJCNN}, 2020.

\bibitem{spinemap}
A.~Balaji \emph{et~al.}, ``{Mapping spiking neural networks to neuromorphic
  hardware},'' \emph{TVLSI}, 2020.

\bibitem{das2018mapping}
A.~Das \emph{et~al.}, ``Mapping of local and global synapses on spiking
  neuromorphic hardware,'' in \emph{DATE}, 2018.

\bibitem{frameworkCAL}
A.~Balaji \emph{et~al.}, ``{A framework to explore workload-specific
  performance and lifetime trade-offs in neuromorphic computing},'' \emph{CAL},
  2019.

\bibitem{reneu}
S.~Song \emph{et~al.}, ``{Improving dependability of neuromorphic computing
  with non-volatile memory},'' in \emph{EDCC}, 2020.

\bibitem{NeuromorphicLR}
S.~Song \emph{et~al.}, ``A case for lifetime reliability-aware neuromorphic
  computing,'' in \emph{MWSCAS}, 2020.

\bibitem{lcpc}
T.~Titirsha \emph{et~al.}, ``Thermal-aware compilation of spiking neural
  networks to neuromorphic hardware,'' in \emph{LCPC}, 2020.

\bibitem{balaji2020run}
A.~Balaji \emph{et~al.}, ``Run-time mapping of spiking neural networks to
  neuromorphic hardware,'' \emph{JSPS}, 2020.

\bibitem{ovshinsky1968reversible}
S.~Ovshinsky, ``Reversible electrical switching phenomena in disordered
  structures,'' \emph{Physical Review Letters}, 1968.

\bibitem{LeeISCA2009}
B.~C. Lee \emph{et~al.}, ``Architecting phase change memory as a scalable dram
  alternative,'' in \emph{ISCA}, 2009.

\bibitem{QureshiISCA09}
M.~K. Qureshi \emph{et~al.}, ``Scalable high performance main memory system
  using phase-change memory technology,'' in \emph{ISCA}, 2009.

\bibitem{palp}
S.~Song \emph{et~al.}, ``Enabling and exploiting partition-level parallelism
  (palp) in phase change memories,'' \emph{TECS}, 2019.

\bibitem{mneme}
S.~Song \emph{et~al.}, ``{Exploiting inter-and intra-memory asymmetries for
  data mapping in hybrid tiered-memories},'' in \emph{ISMM}, 2020.

\bibitem{datacon}
S.~Song \emph{et~al.}, ``{Improving phase change memory performance with data
  content aware access},'' in \emph{ISMM}, 2020.

\bibitem{hebe}
S.~Song \emph{et~al.}, ``{Aging Aware Request Scheduling for Non-Volatile Main
  Memory},'' in \emph{ASP-DAC}, 2021.

\bibitem{kheradpisheh2018stdp}
S.~R. Kheradpisheh \emph{et~al.}, ``{STDP-based spiking deep convolutional
  neural networks for object recognition},'' \emph{Neural Networks}, 2018.

\bibitem{fei2006one}
L.~Fei-Fei \emph{et~al.}, ``One-shot learning of object categories,''
  \emph{TPAMI}, 2006.

\bibitem{allred2019stimulating}
J.~M. Allred \emph{et~al.}, ``Stimulating stdp to exploit locality for lifelong
  learning without catastrophic forgetting,'' Purdue University West Lafayette
  United States, Tech. Rep., 2019.

\bibitem{shim2017biologically}
M.~S. Shim \emph{et~al.}, ``Biologically inspired reinforcement learning for
  mobile robot collision avoidance,'' in \emph{IJCNN}, 2017.

\bibitem{Avrami1941}
M.~Avrami, ``{Granulation, phase change, and microstructure kinetics of phase
  change. III},'' \emph{The Journal of Chemical Physics}, 1941.

\bibitem{Strukov2016}
D.~B. Strukov, ``{Endurance-write-speed tradeoffs in nonvolatile memories},''
  \emph{Applied Physics A: Materials Science and Processing}, 2016.

\bibitem{kennedy2010particle}
J.~Kennedy, ``Particle swarm optimization,'' \emph{Encyclopedia of machine
  learning}, 2010.

\bibitem{lenet}
Y.~LeCun \emph{et~al.}, ``Lenet-5, convolutional neural networks,'' \emph{URL:
  http://yann. lecun. com/exdb/lenet}, 2015.

\bibitem{deng2012mnist}
L.~Deng, ``The mnist database of handwritten digit images for machine learning
  research,'' \emph{IEEE Signal Processing Magazine}, 2012.

\bibitem{alexnet}
A.~Krizhevsky \emph{et~al.}, ``Imagenet classification with deep convolutional
  neural networks,'' in \emph{Advances in neural information processing systems
  (NeurIPS)}, 2012.

\bibitem{deng2009imagenet}
J.~Deng \emph{et~al.}, ``Imagenet: A large-scale hierarchical image database,''
  in \emph{Conference on Computer Vision and Pattern Recognition (CVPR)}, 2009.

\bibitem{vgg16}
K.~Simonyan \emph{et~al.}, ``Very deep convolutional networks for large-scale
  image recognition,'' \emph{arXiv}, 2014.

\bibitem{das2018heartbeat}
A.~Das \emph{et~al.}, ``Heartbeat classification in wearables using multi-layer
  perceptron and time-frequency joint distribution of ecg,'' in \emph{CHASE},
  2018.

\bibitem{HeartClassJolpe}
A.~Balaji \emph{et~al.}, ``{Power-accuracy trade-offs for heartbeat
  classification on neural networks hardware},'' \emph{JOLPE}, 2018.

\bibitem{moody2001physionet}
G.~B. Moody \emph{et~al.}, ``Physionet: a web-based resource for the study of
  physiologic signals,'' \emph{IEEE Engineering in Medicine and Biology
  Magazine}, 2001.

\bibitem{Diehl2015}
P.~U. Diehl \emph{et~al.}, ``{Unsupervised learning of digit recognition using
  spike-timing-dependent plasticity},'' \emph{Frontiers in Computational
  Neuroscience}, 2015.

\bibitem{carlsim}
T.~Chou \emph{et~al.}, ``{CARLsim 4: An open source library for large scale,
  biologically detailed spiking neural network simulation using heterogeneous
  clusters},'' in \emph{International Joint Conference on Neural Networks
  (IJCNN)}, 2018.

\bibitem{HeartEstmNN}
A.~Das \emph{et~al.}, ``{Unsupervised heart-rate estimation in wearables with
  Liquid states and a probabilistic readout},'' \emph{Neural Networks}, 2018.

\bibitem{Kashyap2018}
H.~J. Kashyap \emph{et~al.}, ``{A recurrent neural network based model of
  predictive smooth pursuit eye movement in primates},'' in \emph{International
  Joint Conference on Neural Networks (IJCNN)}, 2018.

\bibitem{dan2004spike}
Y.~Dan \emph{et~al.}, ``Spike timing-dependent plasticity of neural circuits,''
  \emph{Neuron}, vol.~44, pp. 23--30, 2004.

\end{thebibliography}
